\documentclass{article}
\usepackage{ijcai22}

\usepackage{times}

\usepackage{soul}
\usepackage{url}
\usepackage{array}
\usepackage[hidelinks]{hyperref}
\usepackage[utf8]{inputenc}
\usepackage[small]{caption}
\usepackage{graphicx}
\usepackage{amsmath}
\usepackage{booktabs}
\usepackage{float}
\usepackage{algpseudocode}
\usepackage{todonotes}
\usepackage[linesnumbered,ruled,vlined]{algorithm2e}
\usepackage{latexsym} 
\usepackage{amssymb}
\usepackage{amsthm}
\usepackage{makecell}
\usepackage{subfigure}
\usepackage{booktabs}
\usepackage{bm}
\usepackage[normalem]{ulem}
\usepackage{multirow}
\usepackage{multicol}

\newcommand{\el}{$\mathcal{EL}^{++}$}
\renewcommand{\Re}{\mathbb{R}}

\usepackage{cleveref}

\newlength\myindent
\title{Description Logic \el Embeddings with Intersectional Closure}

\author{
Xi Peng\and
Zhenwei Tang\and
Maxat Kulmanov\and
Kexin Niu\and
Robert Hoehndorf\footnote{Contact Author}\\
\affiliations
 Computer, Electrical and Mathematical Sciences \& Engineering Division (CEMSE), Computational Bioscience Research Center (CBRC), King Abdullah University of Science and Technology, Thuwal 23955, Saudi Arabia
\emails
\{xi.peng, zhenwei.tang, maxat.kulmanov, kexin.niu, robert.hoehndor \}@kaust.edu.sa
}

\begin{document}

\maketitle

\begin{abstract}
  Many ontologies, in particular in the biomedical domain, are based
  on the Description Logic \el. Several efforts have been made to
  interpret and exploit \el ontologies by distributed representation
  learning. Specifically, concepts within \el theories have been
  represented as $n$-balls within an $n$-dimensional embedding
  space. However, the intersectional closure is not satisfied when
  using $n$-balls to represent concepts because the intersection of
  two $n$-balls is not an $n$-ball. This leads to challenges when
  measuring the distance between concepts and inferring equivalence
  between concepts. To this end, we developed \underline{EL}
  \underline{B}ox \underline{E}mbedding (ELBE) to learn Description
  Logic \el embeddings using axis-parallel boxes.  We generate
  specially designed box-based geometric constraints from \el axioms
  for model training. Since the intersection of boxes remains as a box, the intersectional
  closure is satisfied. We report extensive experimental results on
  three datasets and present a case study to demonstrate the
  effectiveness of the proposed method.
\end{abstract}

\section{Introduction}
\el is a lightweight Description Logic that admits sound and complete
reasoning in polynomial time and has been used to define the Web
Ontology Language (OWL) 2 EL profile \cite{motik2009owl}. \el is of
great value in real-world applications, and especially in the life
sciences where a large number of ontologies have been developed
\cite{smith2007obo} and the OWL 2 EL profile is widely used
\cite{hoehndorf2011common}.

Statistical methods \cite{subramanian2005gene} and semantic similarity
measures \cite{pesquita2009semantic} have been developed and widely
used to analyze ontologies as well as entities characterized with
concepts in ontologies. Recent years have witnessed increasing
interest in distributed representation learning in a variety of fields
including natural language processing \cite{mikolov2013distributed}
and recommender systems \cite{he2017neural}. This motivated efforts in
understanding and exploiting the Description Logic \el through
embeddings \cite{kulmanov2019embeddings,mondala2021emel}.
The embedding methods for \el consider concepts as $n$-balls in an
$n$-dimensional embedding space. The rationality behind representing
concepts as geometric regions rather than single points in the
embedding space is twofold. First, logical operations within
ontological axioms, such as subsumptions and intersections, require
geometric operations on top of regional representations. Second, the
semantics of \el interprets concepts as sets of entities, and these
are naturally represented as regions in the embedding
space. Therefore, geometric embeddings not only fulfill the need of
embedding \el axioms but also establish a direct correspondence to the
semantics of the \el Description Logic.

\begin{figure}[t]
	\centering  
	\includegraphics[width=0.3\textwidth,trim=0 340 650 0,clip]{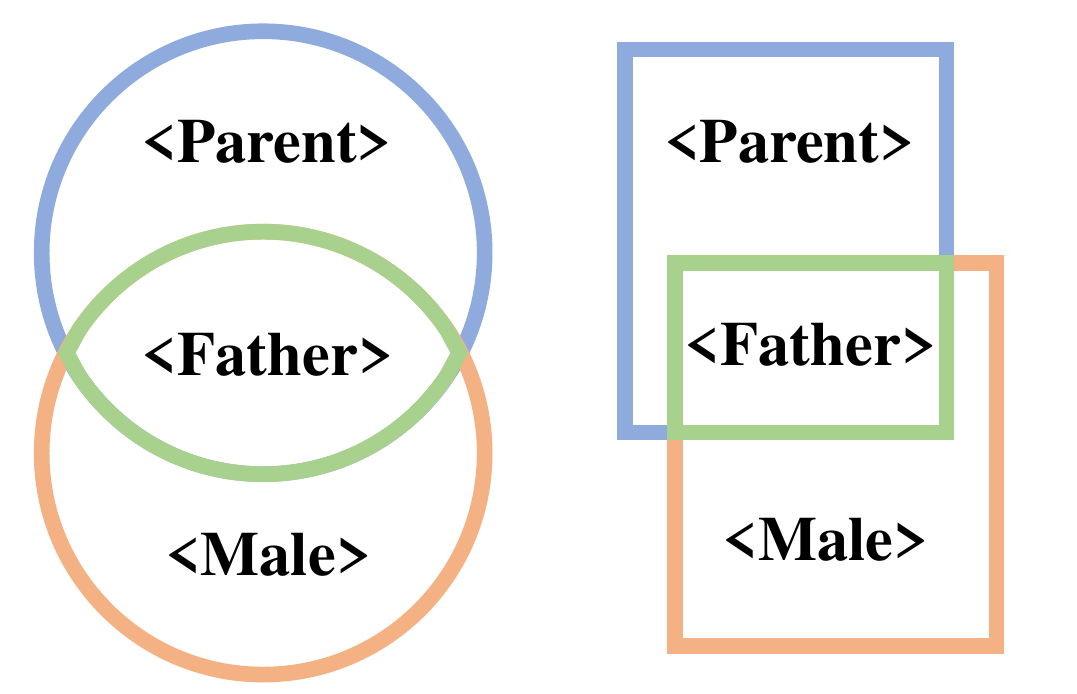}
	\vspace{-0.2cm}
	\caption{Example of concept intersections in $2$-dimensional
          space. The intersection of two boxes remains as a
          box whereas the intersection of $2$-balls is a
          lens-shaped region and not a $2$-ball.}
	\label{idea}
\vspace{-0.3cm}
\end{figure}

Although these embedding methods are effective and theoretically
well-motivated due to the relation between geometric regions and
interpretations of the axioms, a common key issue remains unsolved:
the \textbf{intersectional closure}, i.e., the property that the
intersection of two concepts is again a concept within the embedding
space, is not satisfied. Intuitively, the intersection of concept
$Parent$ and concept $Male$ should also be a concept (i.e.., the
concept $Father$) with the same type of geometric
representation. However, as shown on the left of Figure \ref{idea},
the intersection of two $n$-balls is no longer an $n$-ball. Without
the property of intersectional closure, conflicts arise when
generating embeddings; for example, if the ontology contains both the
axioms $Parent \sqcap Male \sqsubseteq Father$ and
$Father \sqsubseteq Male \sqcap Parent$, the only valid solution to
finding an embedding is to make $Father$, $Male$, and $Parent$
co-extensional; if additional axioms require $Parent$ to contain
entities that are disjoint with $Male$ (such as for $Mother$), no
valid embedding for the axioms can be found. This leads to instability
during learning and challenges in determining similarity between
concepts in embedding space, and it breaks the property that
generating an embedding should generate a model of the theory.

To this end, we developed \underline{EL} \underline{B}ox
\underline{E}mbeddings (ELBE) to learn \el embeddings that satisfy the
\textbf{intersectional closure}. As pointed out by works on geometric
embeddings in other research fields \cite{ren2019query2box}, boxes,
i.e., the area contained within axis-parallel high-dimensional
rectangles, are particularly suitable for modeling logical operations
in the embedding space. As shown on the right of Figure \ref{idea},
the intersection of $2$-dimensional boxes remains as a $2$-dimensional
box and therefore satisfies the intersectional closure; the same
property holds in higher dimensions. Therefore, we propose to use
boxes as the geometric representations of concepts in \el based
ontologies. To do so, we specifically design box-based constraints
that embed axioms of \el ontologies.
Our main contributions are summarized as below:
\begin{itemize}
\item We propose a novel method named ELBE that uses axis-parallel
  boxes to satisfy intersectional closure for \el
  embeddings;
\item We conduct extensive experimental evaluations and provide
  in-depth analysis on real-world benchmark datasets to demonstrate
  the effectiveness the proposed method;
\item We present a case study on a synthetic dataset to gain insights
  of our proposed method.
\end{itemize}

\section{Related work}

\subsection{Description Logic \el}
\el falls into the family of description logics, which are fragments
of first order logic and vary in their expressiveness.
%
%
\el is particular suitable for representing and reasoning over large
ontologies due to its polynomial time complexity in inferring
subsumptions. Therefore, many large ontologies, especially in the
biomedical domain, are based on \el, including the Gene Ontology
\cite{ashburner2000gene}, Human Phenotype Ontology
\cite{kohler2017human}, and SNOMED CT
\cite{donnelly2006snomed}. Furthermore, \el forms the basis of the OWL
2 EL profile \cite{motik2009owl}.

\subsection{Distributed Representation Learning for \el}
Recently, inspired by the effectiveness of distributed representation
learning in a variety of fields
\cite{mikolov2013distributed,he2017neural}, several works propose to
learn embeddings for concepts in \el ontologies.  Specifically,
\cite{kulmanov2019embeddings,mondala2021emel} learn
\textit{geometric} embeddings for \el ontologies.  The key idea of
learning geometric embeddings is that the embedding function projects
the symbols used to formalize \el axioms into an interpretation
$\mathcal{I}$ of these symbols such that $\mathcal{I}$ is a model of
the \el ontology. Other approaches
\cite{chen2021owl2vec,smaili2019opa2vec} rely on more \textit{general}
graph embeddings or word embeddings and apply them to ontology axioms.

In this work, we learn \el embeddings using axis-parallel boxes, which
falls into the \textit{geometric} embedding based methods. The aim of
ELBE is to generate a model for a theory where the points that are
contains within an $n$-box are the extension of a concept. By ensuring
intersectional closure, our novel method addresses the instability in
the learning process and the difficulty of similarity measurement in
previous works. Furthermore, it can represent the extension of concept
intersections within the embedding space.


\subsection{Knowledge Graph Embedding}
Distributed representation learning on knowledge graphs, i.e.,
knowledge graph embedding, has been extensively explored during recent
years. Representative methods include methods that combine random
walks with a language model
\cite{grover2016node2vec,ristoski2016rdf2vec}, translation based
methods \cite{bordes2013translating,wang2014knowledge} that regard a
relation as the translation from a head entity to a tail entity,
tensor decomposition methods
\cite{yang2014embedding,balavzevic2019tucker} that assume the score of
a triple can be decomposed into several tensors, and deep learning
methods \cite{cai2018kbgan,vashishth2020compositionbased} that utilize
deep neural networks to embed knowledge graphs. Although we follow the
well-established translation based knowledge graph embedding method
TransE \cite{bordes2013translating} in terms of the geometric
interpretation of relational mappings, there is a key difference to be
noted: these methods embed entities and relations in knowledge graphs
from triples, while we embed concepts and relations in ontologies
based on \el axioms.

\section{Preliminaries}
In this section, we first formulate the problem we aim to solve. Then,
we introduce the key terminologies in description logic \el and
elaborate the limitations of previous geometric embedding methods for
\el ontologies.


\subsection{Problem Formulation}
An ontology in the Description Logic \el is formulated as
$\mathcal{O}=(C,R,I; ax)$ where $C$ is a set of concept symbols, $R$ a
set of relation symbols, $I$ a set of individual symbols, and $ax$ a
set of axioms.
We aim to find an embedding $e: \mathcal{O} \mapsto \Re^n$ such that
the image of $e$ is a model of $\mathcal{O}$. Furthermore, the
the embedding should allow answering queries: given a query concept
description $Q$, find all concepts $C$ such that $C \sqsubseteq Q$;
more specifically, we rank concepts $C$ to find the top-$k$ concepts
that satisfy the query.


\subsection{\el Terminologies}
\label{elterminologies}
In \el, the TBox, i.e., terminological box, contains axioms describing
concept hierarchies, while the ABox, i.e., assertional box, contains
axioms stating the relations between individuals (or entities) and
concepts. The ABox axioms can be eliminated by replacing $C(a)$ with
$\{a\} \sqsubseteq C$ and replacing $r(a,b)$ with
$\{a\}\sqsubseteq \exists r.{b}$. The Tbox axioms in \el can be
normalized into one of the seven normal forms (NFs) summarized in
Table \ref{nfs}.  The syntax and semantics of \el is summarized in
Table \ref{tbl:el}.

\begin{table}[htbp]\centering
    \renewcommand\arraystretch{1.0}
  	\renewcommand\tabcolsep{10pt}
        \caption{\label{nfs}Normal forms of \el and their
          abbreviations. }
  \begin{tabular}{c|c}
   \toprule
    \textbf{Abbr.} & \textbf{TBox Axioms}  \\ 
    \midrule
    NF1 & $C$ $\sqsubseteq$ $D$ \\
    \midrule
    NF2 & $C\sqcap D \sqsubseteq E$ \\
    \midrule
    NF3 & $\exists R.C \sqsubseteq D$ \\
    NF4 & $C \sqsubseteq \exists R.D$ \\
    \midrule
    NF5 & $C\sqcap D \sqsubseteq \bot$ \\
    NF6 & $\exists R.C \sqsubseteq \bot$ \\
    NF7 & $C\sqsubseteq \bot$ \\
    \bottomrule
  \end{tabular}
\end{table}

\begin{table}[htbp]
\footnotesize
    \renewcommand\arraystretch{1.0}
  	\renewcommand\tabcolsep{5pt}
  \centering
   \caption{\label{tbl:el}Syntax and semantics of \el. }
  \begin{tabular}{c|c|c}
    \toprule
    {\bf Name} & \textbf{Syntax} & \textbf{Semantics} \\
    \midrule
    top & $\top$ & $\Delta^{\mathcal{I}}$ \\
   \midrule
    bottom & $\bot$ & $\emptyset$ \\
    \midrule
    nominal & $\{ a \} $ & $\{ a^{\mathcal{I}} \}$ \\
   \midrule
    conjunction & $C \sqcap D$ & $ C^{\mathcal{I}} \cap
                                 D^{\mathcal{I}}$ \\
   \midrule
   existential
 & \multirow{2}[2]{*}{$\exists r.C$} & $ \{ x \in
                                              \Delta^{\mathcal{I}} |
                                              \exists y \in
                                              \Delta^{\mathcal{I}} :$ \\
                                         restriction   &  & $(x,y) \in
                                              r^{\mathcal{I}} \land y
                                              \in C^{\mathcal{I}} \} $
    \\
   \midrule
    concept inclusion & $C \sqsubseteq D$ &
                                                        $C^{\mathcal{I}}
                                                        \subseteq
                                                        D^{\mathcal{I}}$
    \\
    \midrule
    instantiation & $C(a)$ & $a^\mathcal{I} \in C^\mathcal{I}$ \\
    \midrule
    role assertion & $r(a,b)$ & $(a^\mathcal{I},b^\mathcal{I}) \in r^\mathcal{I}$ \\
    \bottomrule
  \end{tabular}
 
\end{table}

\subsection{Limitations of Previous Works}
Methods that construct geometric models for logical theories based on
$n$-balls, such as ELEm \cite{kulmanov2019embeddings}, have a crucial
limitation. Representing ontology concepts as $n$-balls entails that
they cannot represent intersections of concepts within the same
formalism because intersections of $n$-balls are not $n$-balls. This
does prevent embeddings based on $n$-balls to represent and infer
equivalent concept axioms and model the second normal form
($C \sqcap D \sqsubseteq E$) naturally. Specifically, it leads to
complications with a loss function designed to minimize the loss for
the second normal form, and likely results in embeddings of lower
quality as a consequence.

\section{Methodology}
We consider the seven types of TBox axioms in \el ontologies as
training instances. In this section, we first detail our specially
designed box-based objective functions for each type of the TBox
axioms. Then we present a case study in the family domain to
demonstrate the rationalness of our proposed ELBE.

\begin{figure*}[!t]
\centering
\subfigure[]{\includegraphics[width=2.2in,trim=50 10 200 110,clip]{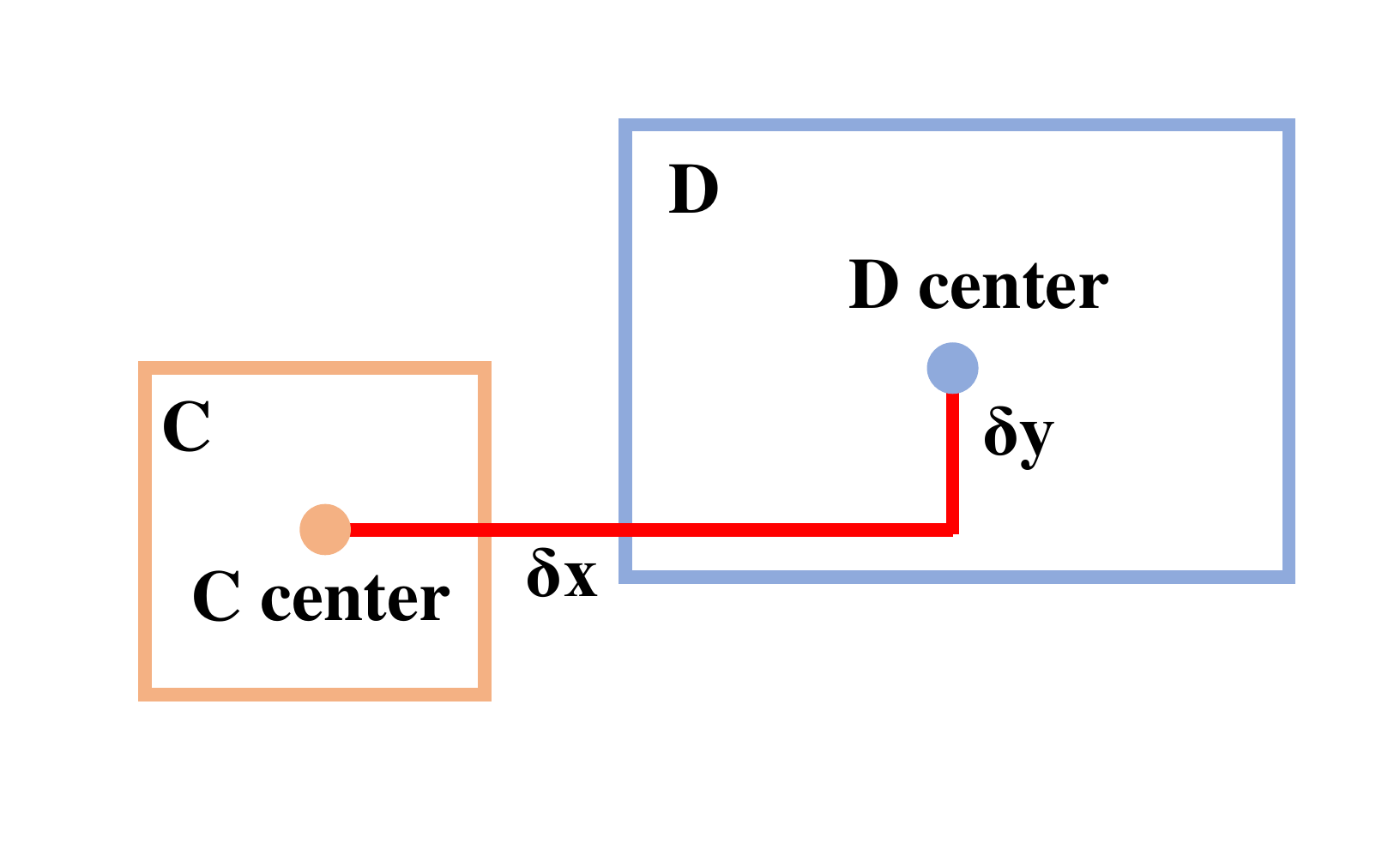}\label{dis1}}
\subfigure[]{\includegraphics[width=2.2in,trim= 0 0 230 90,clip]{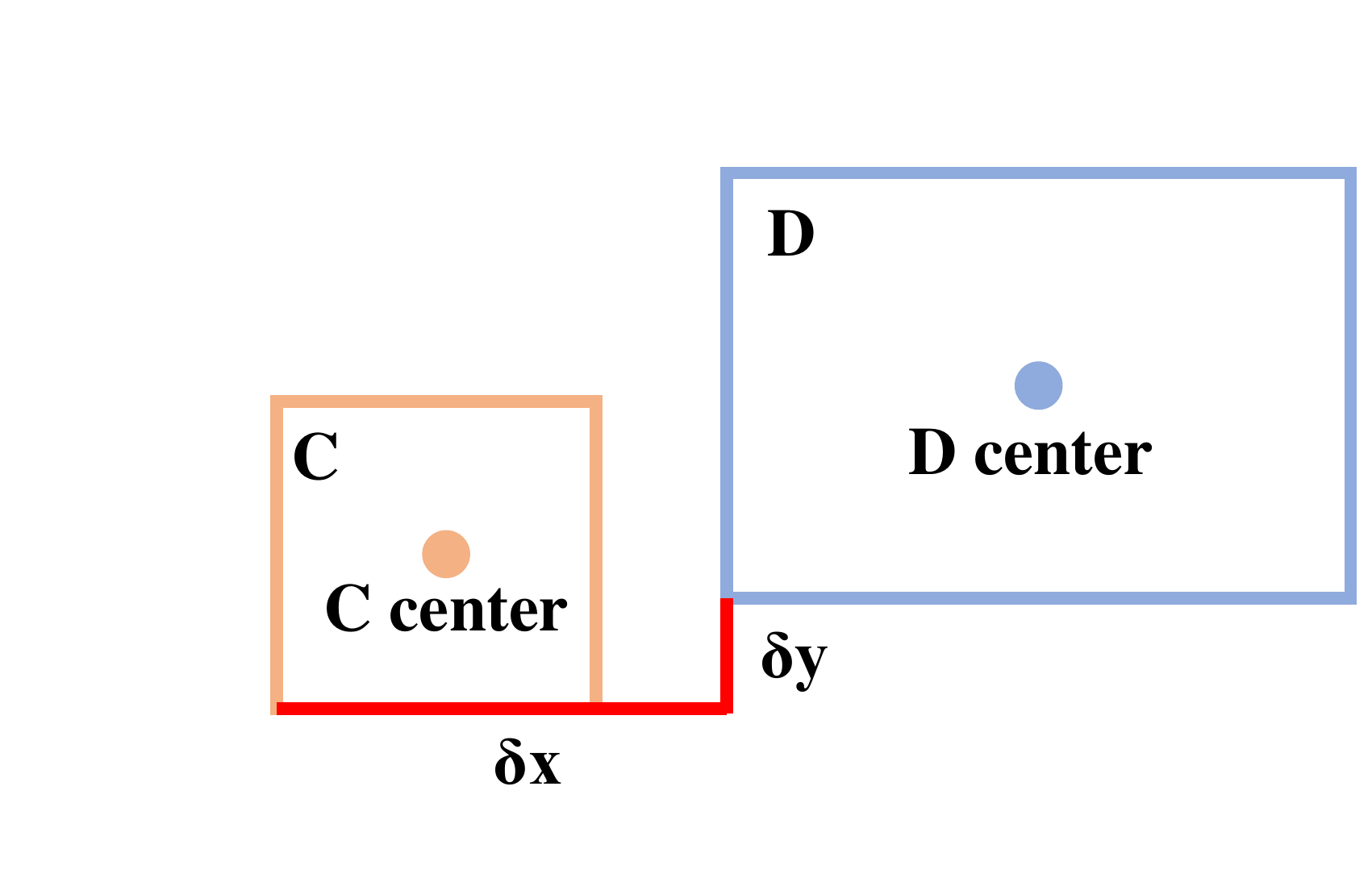}\label{dis2}}
\subfigure[]{\includegraphics[width=2.2in,trim=250 50 0 60,clip]{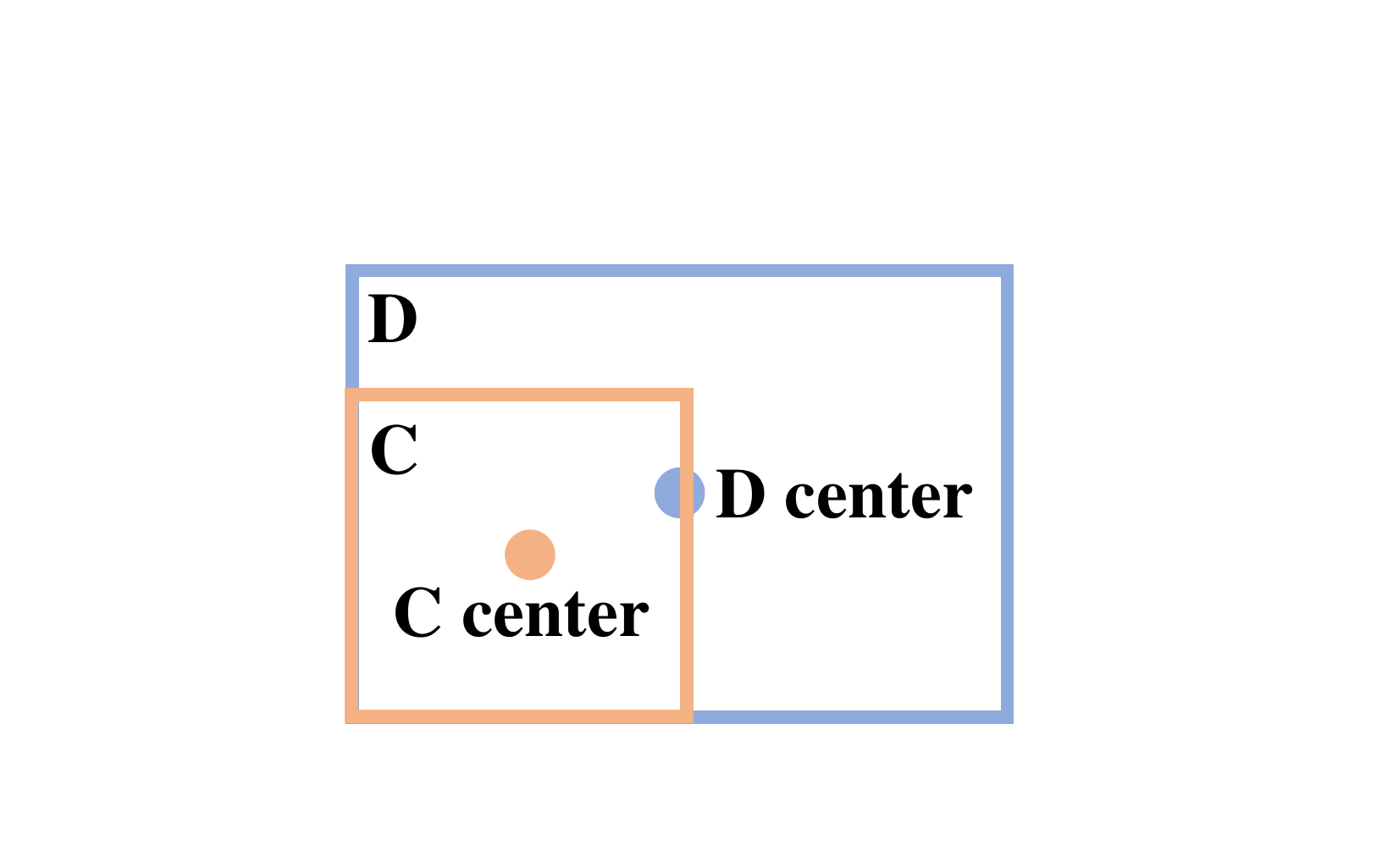}\label{dis3}}
\caption{(a) The distance between centers of two boxes. (b) The distance with offsets. (c) When $\delta x$, $\delta y$ are both 0. }
\label{fig5}
\end{figure*}


\subsection{First Normal Form (NF1)}
We use $\beta (C)$ to represent the box embedding of concept
$C$. Inspired by \cite{ren2020query2box}, we use two vectors to
represent a box. One vector represents the center of the box and the
other vector defines the offset. The center is the intersection of the
diagonals. The offset is the vector that represents the size of each
dimension of the box, so all the elements are non-negative values. Let
$e_c: C \cup R \longrightarrow \mathbb{R}^n$ be the mapping function
that maps each concept to the center of its box embedding and that
maps each relation to its embedding. Let
$e_o: C \longrightarrow \mathbb{R}^n$ be the mapping function that
maps each concept to the n-dimension vector space as the offset of the
box.

The loss function for the first normal form, $C \sqsubseteq D$, aims
to ensure that the box embedding of concept $C$ lies entirely within
the box embedding of concept $D$.
Figure \ref{dis1} illustrates two concepts $C$ and $D$ with their two
box embedding $\beta (C)$ and $\beta (D)$. The red line indicates the
difference between the center vectors:
$| e_c(D)-e_c(C) |=(\delta x, \delta y) $ where $\delta x$ and
$\delta y $ are two non-negative real number. In Figure \ref{dis2}, we
show how the red line defined by $\delta x$ and $\delta y $ is
computed as
$| e_c(D)-e_c(C)| + e_o(C) - e_o(D) =(\delta x, \delta y) $. As shown
in Figure \ref{dis3}, when both $\delta x$, $\delta y$ are $0$ or less
than $0$, then $\beta (D)$ will contain $\beta (C)$. When we extend to
higher dimensions, if the elements of difference vector (in two
dimension is $(\delta x, \delta y)$) are all less or equal to $0$,
then $\beta (C)$ is contained within $\beta (D)$. So the loss function
of the first normal form, $C \sqsubseteq D$, and which ensures that
$\beta(C)$ is contained within $\beta(D)$ is:
\begin{equation}
\begin{split}
  loss_{C \sqsubseteq D}(c,d) =
  &||max(zeros,\left|e_c(c)-e_c(d)\right|\\
  &+e_o(c)-e_o(d)-margin)|| \\
\end{split}
\end{equation}
For all loss functions we use a margin vector. If all elements of
margin vector are no larger than $0$, then $\beta (C)$ lies properly
in $\beta (D)$.

\subsection{Second Normal Form (NF2)}
In the embeddings space, the second norm form
$C\sqcap D \sqsubseteq E$ implies that the intersection of $\beta (C)$
and $\beta (D)$ is contained within $\beta (E)$. The key aim for the
loss is to find the intersection of two boxes:
$e_c(new)=(box_{min}+box_{max})/2 $, and
$e_o(new)=\left|(box_{max}-box_{min})\right|/2$. It will be computed
using the concepts $C$ and $D$ by equation \ref{left} and
\ref{right}. We compute
\begin{equation}
box_{min} = max(e_c(C)-e_o(C),e_c(D)-e_o(D))
\label{left}
\end{equation}
and
\begin{equation}
box_{max} = min(e_c(C)+e_o(C),e_c(D)+e_o(D))
\label{right}
\end{equation}
from which we can obtain the center and offset of the
intersection. For this new embedding representing the intersection of
two concepts, we apply the same loss as for the first normal form
(where concept $C$ from the first norm form is now the intersection of
the concepts $C$ and $D$ in the second norm form). The loss function
for the second normal form $C \sqcap D \sqsubseteq E$ will be:
\begin{equation}
\begin{split}
loss_{C\sqcap D \sqsubseteq E}(c,d,e)=||max(zeros,|e_c(new)\\
-e_c(e)|+e_o(new)-e_o(e)-margin)||
\end{split}
\end{equation} 

\subsection{Normal Forms with $R$ (NF3, NF4)}
The first two normal forms do not include any quantifiers or
relations. Every point that lies properly within a box representing a
concept is an entity that lies in the extension of the concept (see
Table \ref{tbl:el}), and we apply relations as transformations on
these points. We use TransE \cite{bordes2013translating} to represent
the relations between these entities. The losses in Equations
\ref{nf3} and \ref{nf4} capture this intention; in particular, the
extension of the concept $\exists R.D$ is the transformation of all
entities in the extension of $D$ by $-e_c(R)$.
\begin{equation}
\begin{split}
loss_{C \sqsubseteq \exists R.D}(c,d,r) =||max(zeros,|e_c(c)
\\+e_c(r)-e_c(d)|+e_o(c)-e_o(d) -margin)||
\end{split}
\label{nf3}
\end{equation}

\begin{equation}
\begin{split}
loss_{\exists R.C \sqsubseteq D}(c,d,r) =||max(zeros,|e_c(c)
\\-e_c(r)-e_c(d)|-e_o(c)-e_o(d)-margin)||
\end{split}
\label{nf4}
\end{equation}

\begin{figure}[H]
   \centering
  \includegraphics[width=0.5\textwidth,trim=60 400 100 50,clip]{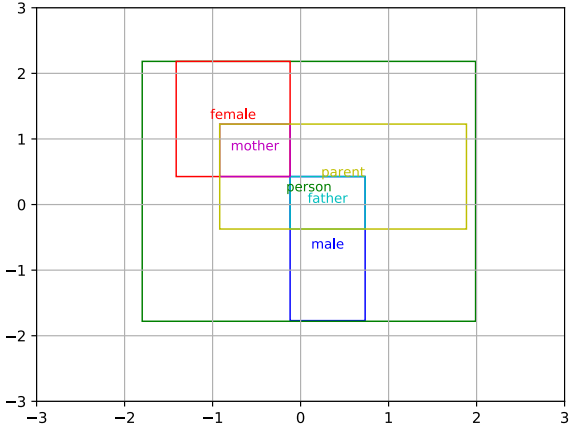}
  \caption{Visualization of embeddings in the family domain.}
\label{family}
\end{figure}

\subsection{Normal Forms with $\bot$ (NF5, NF6, NF7)}
In \el, the $\bot$ concept can only appear on the right-hand side of
three of the normal forms: $C\sqcap D \sqsubseteq \bot$,
$\exists R.C \sqsubseteq \bot$, and $C \sqsubseteq \bot$. Intuitively,
the first normal form states that the concepts $C$ and $D$ are
disjoint. The box embedding of the intersection of $C$ and $D$ can be
calculated by Eqn. \ref{left} and \ref{right}. If at least one element
in $box_{max}$ is less then the corresponding element in $box_{min}$,
then there is no intersection and the loss is zero; otherwise, the
loss is as Eqn. \ref{nf2Dis}:
\begin{equation}
\begin{split}
loss_{C\sqcap D \sqsubseteq \bot}(c,d) =
&||max(zeros,-\left|e_c(c)-e_c(d)\right|\\
&+e_o(c)+e_o(d)+margin)||\\
\end{split}
\label{nf2Dis}
\end{equation}
The loss functions in Equations \ref{nf1Dis} and \ref{nf4Dis} capture
the intuition that if a concept is unsatisfiable then its embedding
should have no extensions; we use a very similar loss of the normal
form where $\exists R.C$ is unsatisfiable due to our relation model
based on linear transformations.
\begin{equation}
\begin{split}
loss_{C \sqsubseteq \bot}(c) = ||e_o(c)||
\end{split}
\label{nf1Dis}
\end{equation}
\begin{equation}
\begin{split}
loss_{\exists R.C \sqsubseteq \bot}(c,r) = ||e_o(c)||
\end{split}
\label{nf4Dis}
\end{equation}

We also add negatives to improve the performance of our method:
\begin{equation}
\begin{split}
loss_{C \not\sqsubseteq \exists R.D}(c,d,r) =||max(zeros,-|e_c(c)\\
+e_c(r)-e_c(d)|+e_o(c)+e_o(d)+margin)||
\end{split}
\label{nf3neg}
\end{equation}

\subsection{Case Study: The Family Domain}

We construct a simple knowledge base to test and understand the
embeddings of our model. We use the family domain in which we generate
a knowledge base that contains examples for each of the normal forms
(Eqn. \ref{fam1}--\ref{famlast}). We chose margin $margin={0}$ and an
embedding dimension of $2$ so that we can visualize the generated
embeddings in $\Re^2$. As shown in Figure \ref{family}, ELBE can
correctly model the relationship between each concepts, especially the
intersection of two concepts. In the family domain, we can infer that
$Mother$ is equivalent to the intersection of $Female$ and $Parent$,
and $Father$ is equivalent to the intersection of $Male$ and $Parent$.

\begin{eqnarray}
  \label{fam1}
& Male & \sqsubseteq Person \\
& Female & \sqsubseteq Person \\
& Father & \sqsubseteq Male \\
& Mother & \sqsubseteq Female \\
& Father & \sqsubseteq Parent \\
& Mother & \sqsubseteq Parent \\
& Female \sqcap Male & \sqsubseteq \bot \\
& Female \sqcap Parent & \sqsubseteq Mother \\
& Male \sqcap Parent & \sqsubseteq Father \\
& \exists hasChild.Person & \sqsubseteq Parent \\
& Parent & \sqsubseteq Person \\
& Parent & \sqsubseteq \exists hasChild.\top
  \label{famlast}
\end{eqnarray}

\section{Experiments}
In this section, we conduct extensive experiments to compare ELBE with
state of the art methods by answering the following research
questions:
\begin{itemize}
\item \textbf{RQ1}: Does the proposed box based method ELBE perform
  better than the state-of-the-art methods?
\item \textbf{RQ2}: How does satisfying intersectional closure
  contribute to learning \el embeddings?
\end{itemize}

\begin{table*}[htbp]
    \renewcommand\arraystretch{1.35}
  	\renewcommand\tabcolsep{5pt}
  \centering
   \begin{tabular}{l|r|r|r|r|r|r|r|r}
    \toprule
    \makecell[c]{Method} & \multicolumn{1}{l|}{H@10(R)} & \multicolumn{1}{l|}{H@10(F)} & \multicolumn{1}{l|}{H@100(R)} & \multicolumn{1}{l|}{H@100(F)} & \multicolumn{1}{l|}{MR(R)} & \multicolumn{1}{l|}{MR(F)} & \multicolumn{1}{l|}{AUC(R)} & \multicolumn{1}{l}{AUC(F)} \\
    \midrule
    \makecell[c]{TransE(R)} & \makecell[c]{0.03}  & \makecell[c]{0.05}  & \makecell[c]{0.22}  & \makecell[c]{0.27}  & \makecell[c]{855}   & \makecell[c]{809}   & \makecell[c]{0.84}  & \makecell[c]{0.85} \\
    
   \makecell[c]{TransE(P)} & \makecell[c]{0.06}  & \makecell[c]{0.13} & \makecell[c]{0.41}  & \makecell[c]{0.54} & \makecell[c]{378}   & \makecell[c]{330} & \makecell[c]{0.93}  & \makecell[c]{0.94} \\
   
    \makecell[c]{SimResnik} & \makecell[c]{0.08}  & \makecell[c]{0.18} & \makecell[c]{0.38}  & \makecell[c]{0.49} & \makecell[c]{713}   & \makecell[c]{663} & \makecell[c]{0.87}  & \makecell[c]{0.88} \\
    
    \makecell[c]{SimLin} & \makecell[c]{0.08}  & \makecell[c]{0.17}  & \makecell[c]{0.34}  & \makecell[c]{0.45}  & \makecell[c]{807}   & \makecell[c]{756}   & \makecell[c]{0.85}  & \makecell[c]{0.86} \\
    

    \makecell[c]{EmEL++} & \makecell[c]{0.07}  & \makecell[c]{0.17}  & \makecell[c]{0.48}  & \makecell[c]{0.65}  & \makecell[c]{336}   & \makecell[c]{291}   & \makecell[c]{0.94}  & \makecell[c]{0.95} \\
    \makecell[c]{ELEm}  & \makecell[c]{\underline{0.10}}   & \makecell[c]{\underline{0.23}}  & \makecell[c]{0.50}   & \makecell[c]{\underline{0.75}}  & \makecell[c]{\underline{247}}   & \makecell[c]{\underline{187}}   & \makecell[c]{\underline{0.96}}  & \makecell[c]{\underline{0.97}} \\
    \midrule
    \makecell[c]{\textbf{ELBE}}  & \makecell[c]{\textbf{0.11}} & \makecell[c]{\textbf{0.26}} & \makecell[c]{\textbf{0.57}} & \makecell[c]{\textbf{0.77}} & \makecell[c]{\textbf{201}} & \makecell[c]{\textbf{154}} & \makecell[c]{\textbf{0.96}} & \makecell[c]{\textbf{0.97}} \\
    \bottomrule
    \end{tabular}%
    \caption{\label{tbl:results-yeast} Prediction performance for yeast
    protein--protein interactions. }
    
\end{table*}%

\begin{table*}[htbp]
    \renewcommand\arraystretch{1.35}
  	\renewcommand\tabcolsep{5pt}
  \centering
    \begin{tabular}{l|r|r|r|r|r|r|r|r}
    \toprule
    \makecell[c]{Method} & \multicolumn{1}{l|}{\makecell[c]{H@10(R)}} & \multicolumn{1}{l|}{\makecell[c]{H@10(F)}} & \multicolumn{1}{l|}{\makecell[c]{H@100(R)}} & \multicolumn{1}{l|}{\makecell[c]{H@100(F})} & \multicolumn{1}{l|}{\makecell[c]{MR(R)}} & \multicolumn{1}{l|}{\makecell[c]{MR(F)}} & \multicolumn{1}{l|}{\makecell[c]{AUC(R)}} & \multicolumn{1}{l}{\makecell[c]{AUC(F)}} \\
    \midrule
    \makecell[c]{TransE(R)} & \makecell[c]{0.02}  & \makecell[c]{0.03}  & \makecell[c]{0.12}  & \makecell[c]{0.16}  & \makecell[c]{2262}  & \makecell[c]{2189}  & \makecell[c]{0.85}  & \makecell[c]{0.85} \\
    
    \makecell[c]{TransE(P)} & \makecell[c]{0.05}  & \makecell[c]{0.11}  & \makecell[c]{0.32} & \makecell[c]{0.44}  & \makecell[c]{809}   & \makecell[c]{737}   & \makecell[c]{0.95}  & \makecell[c]{0.95} \\
    
    \makecell[c]{SimResnik} & \makecell[c]{0.05}  & \makecell[c]{0.10}   & \makecell[c]{0.23}  & \makecell[c]{0.28}  & \makecell[c]{2549}  & \makecell[c]{2476}  & \makecell[c]{0.83}  & \makecell[c]{0.83} \\
    
    \makecell[c]{SimLin} & \makecell[c]{0.04}  & \makecell[c]{0.08}  & \makecell[c]{0.19}  & \makecell[c]{0.22}  & \makecell[c]{2818}  & \makecell[c]{2743}  & \makecell[c]{0.81}  & \makecell[c]{0.82} \\
    

    \makecell[c]{EmEL++} & \makecell[c]{0.04}  & \makecell[c]{0.13}  & \makecell[c]{0.38}  & \makecell[c]{0.56}  & \makecell[c]{772}   & \makecell[c]{700}   & \makecell[c]{0.95}  & \makecell[c]{0.95} \\
    
    \makecell[c]{ELEm}  & \makecell[c]{\underline{0.09}} & \makecell[c]{\underline{0.22}} & \makecell[c]{\underline{0.43}}  & \makecell[c]{\underline{0.70}}   & \makecell[c]{\underline{658}}   & \makecell[c]{\underline{572}}   & \makecell[c]{\underline{0.96}}  & \makecell[c]{\underline{0.96}} \\
    
    \midrule
    \makecell[c]{\textbf{ELBE}}  & \makecell[c]{\textbf{0.09}} & \makecell[c]{\textbf{0.22}} & \makecell[c]{\textbf{0.49}} & \makecell[c]{\textbf{0.72}} & \makecell[c]{\textbf{434}} & \makecell[c]{\textbf{362}} & \makecell[c]{\textbf{0.97}} & \makecell[c]{\textbf{0.98}} \\
    \bottomrule
    \end{tabular}%
    \caption{\label{tbl:results-human} Prediction performance for human
    protein--protein interactions.}
    
\end{table*}%

\subsection{Experimental settings}
\paragraph{Datasets.}
Following the well-established previous works
\cite{kulmanov2019embeddings}, we use two real-world benchmark
datasets to evaluate ELBE; the datasets are used to predict
protein--protein interactions (PPIs) in yeast and humans.
Each protein is associated with its biological functions as expressed
using the Gene Ontology (GO) \cite{ashburner2000gene}, and
interactions are predicted based on the biological hypothesis that
proteins that are functionally similar are more likely to interact.
We use the OWL representation of the datasets where proteins are
instances, and if protein $P$ is associated with the function $F$, we
add the axiom $\{ P \} \sqsubseteq \exists hasFunction.F$ (based on
the ABox axiom $(\exists hasFunction.F)(P)$).  We use 80\% of the
total interacted protein pairs for model training, 10\% and 10\% for
validation and testing, respectively.

In addition, we generate a synthetic dataset to evaluate the
performance of ELBE on the inference of equivalence concept
axioms. The details of this task are in Section \ref{secEEC}. For
generating the dataset, we choose the GO as the basic dataset, then
randomly choose 2,131 axioms of NF2 ($C \sqcap D \sqsubseteq
E$). Then, for each NF2 axiom we chose, we add the axioms
$C \sqsubseteq E$ and $D \sqsubseteq E$ axioms to the dataset. The two
NF1 axioms and one NF2 axiom form a triple that we use for the
entailment of equivalence concepts. We randomly choose 1,000 of these
triples for evaluation and use the rest for training the embeddings.

\paragraph{Baselines}
We compare ELBE with TransE \cite{bordes2013translating}, Resnik's
similarity \cite{resnik1995using}, Lin's similarity
\cite{harispe2015semantic}, EmEL++ \cite{mondala2021emel}, ELEm
\cite{kulmanov2019embeddings}. For TransE, we use two representations,
a native RDF-based rendering of the OWL knowledge base, and a
``plain'' representation based on OWL2Vec* \cite{chen2021owl2vec}
transformation rules, to generate knowledge graph embeddings and use
them for link prediction. For Resnik's similarity and Lin's
similarity, we use the best-match average strategy for combining
pairwise class similarities, then compute the similarity between each
protein. For EmEL++ and ELEm, we predict whether axioms of the type
${P_1} \sqsubseteq \exists interacts.{P_2}$ hold.

\paragraph{Implementation Details.}
We implement ELBE using the Python library
PyTorch\footnote{https://pytorch.org/} and conduct all the experiments
on a Linux server with GPUs (Nvidia RTX 2080Ti) and Intel
Xeon CPU. In the training phase, the initial learning rate of the Adam
\cite{kingma2014adam} optimizer is tuned by grid searching within
\{$1e^{-2}$, $5e^{-3}$, $1e^{-3}$, $5e^{-4}$\} for both tasks. We
perform an extensive search for optimal parameters, testing embedding
sizes for \{$25$, $50$, $100$, $200$, $400$\} and margin vectors for
\{$-0.01$, $-0.05$, $-0.01$, $0$, $0.01$, $0.05$, $0.1$\}. The optimal
set of key hyper-parameters for ELBE is $embedding\_size=50$,
$margin={-0.05}$.

For the PPI task, we evaluate the performance based on hit rate at
ranks 10 (H@10) and 100 (H@100), mean rank (MR) and area under the ROC
curve (AUC) and report both raw and filtered results (e.g., MR(F)
indicates mean rank of the filtered result). For entailment of
equivalent concepts, we evaluate the performance based on hit rate at
ranks 1, 3, and 10, and the mean rank. You can see our code at github\footnote{https://github.com/bio-ontology-research-group/EL2Box\_embedding}.

\subsection{Protein--Protein Interactions (RQ1)}
We use the similarity-based function in Eqn \ref{sim1} for predicting PPIs.
 \begin{equation}
\begin{split}
\label{sim1}
sim(P_1,interact,P_2) =
-||max(zeros,|e_c(P_1)
\\+e_c(interact)-e_c(P_2) |- e_o(P_1)-e_o(P_2))||
\end{split}
\end{equation}
For a query $P_1 \sqsubseteq \exists interacts.P_2$, we predict
interactions of $P_2$ to all proteins from our training set and
identify the rank of $P_1$.

We compare the overall performance of ELBE with that of baselines to
answer RQ1. The results are shown in Table \ref{tbl:results-yeast} for
the yeast PPI dataset and in Table \ref{tbl:results-human} for the
human PPI dataset. The results shows that ELBE consistently
outperforms other baselines.

\subsection{Entailment of Equivalence Concepts (RQ2)}
\label{secEEC}
ELBE outperforms other methods in our experiments, and we hypothesize
this is due to the improved representation of concept intersections. We
perform a more thorough analysis of ELBE's ability to represent and
infer intersections. In the knowledge base based on GO, if there are
three axioms of the form $E \sqsubseteq C$,  $E \sqsubseteq D$ and
$C\sqcap D \sqsubseteq E$, then we can infer (deductively) that $C
\sqcap D \equiv E$. We perform an experiment to test whether ELBE is
able to make these inferences within the embedding space.

To the best of our knowledge, the current methods for \el embeddings
are all based on geometric models based on $n$-balls; we used ELEm
\cite{kulmanov2019embeddings} as representative.



To do the entailment of equivalence concepts task, we predict which
concept equals the intersection of two other concepts. Intuitively, we
should calculate the intersection of concepts first and then compare
the intersection to the embedding of each concept, and choose the
closest one. However, ELEm only approximate the intersection and do
not actually represent it; therefore, we use the center of the
smallest $n$-ball containing the intersection as representation of the
concept and choose the concept that has the closest center of its
$n$-ball embedding. The similarity-based function for ELEm is in
equation \ref{elemSim}.
\begin{equation}
\begin{split}
h=\frac{r_{\eta}(C)^2-r_{\eta}(D)^2+||f_{\eta}(D)-f_{\eta}(C)||^2}{2||f_{\eta}(D)-f_{\eta}(C)||}
\end{split}
\end{equation}
\begin{equation}
\begin{split}
k=\frac{h}{||f_{\eta}(D)-f_{\eta}(C)||}
\end{split}
\end{equation}
\begin{equation}
\begin{split}
sim_{ball}(C,D,E)=|f_{\eta}(C)+k(f_{\eta}(D)-f_{\eta}(C))- f_{\eta}(E)|
\end{split}
\label{elemSim}
\end{equation}

In this similarity function, $r_{eta}$ and $f_{eta}$ are the radius
and center of the $n$-ball of the ELEm.  The similarity-based function
for ELBE is in equation \ref{elbeSim}:
\begin{equation}
\begin{split}
sim_{box}(C,D,E)=|(box_{min}+box_{max})/2-e_c(E)|
\end{split}
\label{elbeSim}
\end{equation}
The $box_{min}$, $box_{max}$ vectors can be calculated by Equations
\ref{left} and \ref{right}. We evaluate the predictive performance
based on recall at rank $1$, rank $3$, rank $10$, and mean rank. The
results are shown in Table \ref{tbl:interClosure}. We find that ELBE
outperforms ELEm in all metrics. 
When we consider only rank $1$ concepts as a result of ``reasoning''
within the embedding space, we find that ELBE improves substantially
over ELEm and can perform inferences of equivalence accurately. In the
test triples, ELBE can predict over 87\% of the equivalence axioms
correctly, while ELEm can predict only 71\%. In other word, by
satisfying the intersectional closure, ELBE can perform entailment of
equivalence; this may also help ELBE to improve the general
performance in other types of queries.

\begin{table}[htbp]
    \renewcommand\arraystretch{1.3}
  	\renewcommand\tabcolsep{5pt}
  \centering
       \caption{\label{tbl:interClosure} Performance for entailment of equivalence concepts tasks.}
    \begin{tabular}{l|r|r|r|r}
    \toprule
    \makecell[c]{Method} & \multicolumn{1}{l|}{\makecell[c]{H@1}} & \multicolumn{1}{l|}{\makecell[c]{H@3}} & \multicolumn{1}{l|}{\makecell[c]{H@10}} & \multicolumn{1}{l}{\makecell[c]{MR}} \\
    \midrule
   \makecell[c]{ELEm} & \makecell[c]{0.710} & \makecell[c]{0.896} & \makecell[c]{0.969} & \makecell[c]{3.561} \\
    \makecell[c]{ELBE} & \makecell[c]{\textbf{0.871}} & \makecell[c]{\textbf{0.974}} & \makecell[c]{\textbf{0.985}} & \makecell[c]{\textbf{3.470}} \\
    \bottomrule
    \end{tabular}%
\vspace{-0.3cm}
\end{table}%

\section{Conclusions}
We developed ELBE, an embedding model for \el ontologies based on
$n$-boxes. In experiments with two datasets from the biomedical domain
as well as one synthetic dataset, we demonstrated that ELBE
outperforms other \el embeddings, knowledge graph embeddings, and
semantic similarity measures. When comparing with other geometric
embeddings of \el ontologies, ELBE solves the intersectional closure
problem and therefore performs better in the entailment of equivalent
concepts axioms. 
Solving the intersectional closure problem is a crucial step towards
establishing a correspondence between vector space embeddings and
models of axiomatic theories.


\section{Future Work}
One limitation of ELBE is the use of TransE as model for
relations. TransE can only deal with one-to-one relations, while
one-to-many and many-to-many relations are important for accurately
embedding ontologies. Furthermore, TransE is a linear transformation
model; consequently, some of our loss functions (in particular
$loss_{\exists R.C \sqsubseteq \bot}$) are insufficient to capture the
\el semantics. In future work, we will explore more expressive
relation models for \el embedding to solve these limitations.
\clearpage
\small
\bibliographystyle{named}
\bibliography{ijcai22}

\end{document}